\documentclass[10pt,twocolumn,letterpaper]{article}
\usepackage{wacv}
\usepackage{graphicx}
\usepackage{subfig}
\usepackage{amsmath}
\usepackage{amssymb}
\usepackage{booktabs}
\usepackage{algorithm}
\usepackage{algorithmic}
\usepackage{tabularx}
\usepackage{placeins}
\usepackage{multirow}
\usepackage{float}
\usepackage{fontenc}
\usepackage{textcomp}

\usepackage[capitalize]{cleveref}
\crefname{section}{Sec.}{Secs.}
\Crefname{section}{Section}{Sections}
\Crefname{table}{Table}{Tables}
\crefname{table}{Tab.}{Tabs.}

\begin{document}

\title{B-FPGM: Lightweight Face Detection via Bayesian-Optimized Soft FPGM Pruning
\thanks{This work was supported by the EU Horizon Europe and Horizon 2020 programmes under grant agreements 101070093 vera.ai and 951911 AI4Media, respectively.}}


\author{Nikolaos Kaparinos\\
CERTH-ITI\\
Thessaloniki, Greece, 57001\\
{\tt\small kaparinos@iti.gr}
\and
Vasileios Mezaris\\
CERTH-ITI\\
Thessaloniki, Greece, 57001\\
{\tt\small bmezaris@iti.gr}
}
\maketitle

\begin{abstract}
Face detection is a computer vision application that increasingly demands lightweight models to facilitate deployment on devices with limited computational resources. Neural network pruning is a promising technique that can effectively reduce network size without significantly affecting performance. In this work, we propose a novel face detection pruning pipeline that leverages Filter Pruning via Geometric Median (FPGM) pruning, Soft Filter Pruning (SFP) and Bayesian optimization in order to achieve a superior trade-off between size and performance compared to existing approaches. FPGM pruning is a structured pruning technique that allows pruning the least significant filters in each layer, while SFP iteratively prunes the filters and allows them to be updated in any subsequent training step. Bayesian optimization is employed in order to optimize the pruning rates of each layer, rather than relying on engineering expertise to determine the optimal pruning rates for each layer. In our experiments across all three subsets of the WIDER FACE dataset, our proposed approach B-FPGM consistently outperforms existing ones in balancing model size and performance. All our experiments were applied to EResFD, the currently smallest (in number of parameters) well-performing face detector of the literature; a small ablation study with a second small face detector, EXTD, is also reported. The source code and trained pruned face detection models can be found at: \url{https://github.com/IDTITI/B-FPGM}.
\end{abstract}

\section{Introduction}
\label{sec:intro}

Face detection is a computer vision task with the objective to ascertain whether any faces are present and, if so, to identify their location and size within a given image. While humans are able to perform this task accurately from an early age, computers find it to be significantly more challenging. The difficulties in face detection arise from various factors in the scene, such as changes in lighting, occlusions, pose and facial scale. These variations demand sophisticated algorithms to achieve reliable detection \cite{minaee2021going}. Despite significant advancements in face detection algorithms over the past few decades, achieving accurate and efficient face detection in real-world applications, such as robotics computer vision \cite{renuka2018automatic, vadakkepat2008multimodal}, security\cite{elrefaei2017real} and surveillance\cite{Kumar2019}, continues to be a challenge.

\begin{figure*}[]
    \centering
    \includegraphics[width=0.89\linewidth]{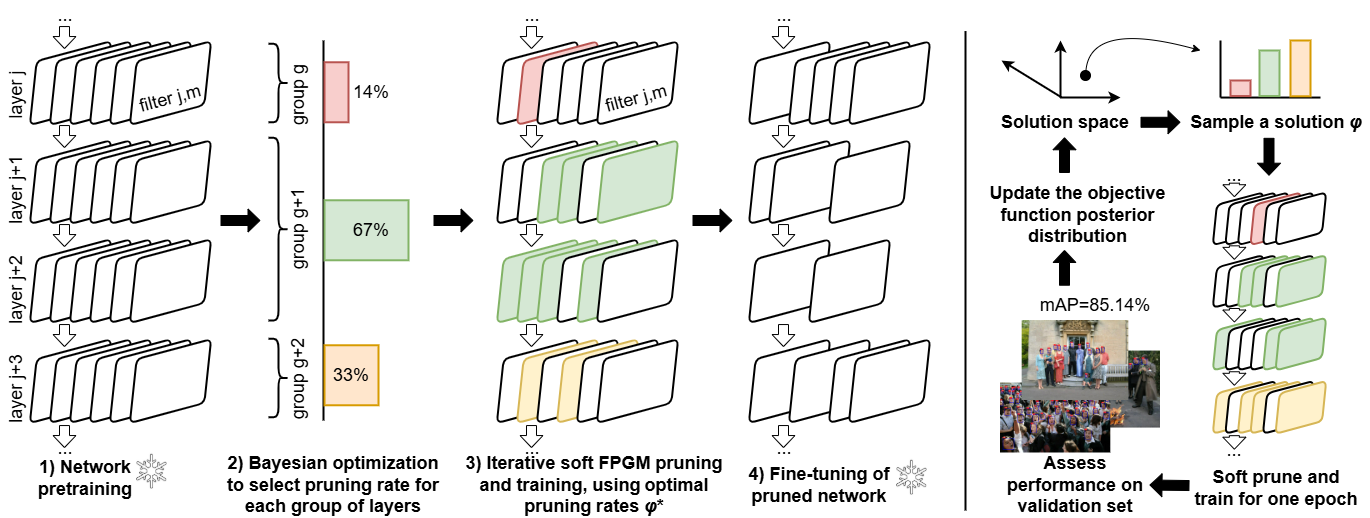}
    \caption{Overview of our proposed pruning and training pipeline. The diagram on the left illustrates our complete methodology, while the diagram on the right elaborates on the iterative Bayesian optimization process (i.e. the 2nd step of the overall pipeline shown on the left). The snowflake symbol indicates that the network's structure remains unchanged during certain steps of the overall pipeline (in contrast to the filter weight values, which are updated throughout all training steps). In contrast to the proposed B-FPGM, the universal FPGM of \cite{gkrispanis2024filter} omits step 2 and prunes all layers using the same pruning rate.}
    \label{fig:overview}
\end{figure*}

Additionally, the increasing deployment of face detectors on edge devices with limited computational resources, such as mobile phones\cite{sarkar2016deep} and drones\cite{rostami2024deep}, has emphasized the need for efficient and compact face detection models. These models must be lightweight enough to run on edge devices without compromising on accuracy. Several solutions have been proposed in the literature, including neural network pruning and knowledge distillation\cite{jin2019learning}.

Pruning is a particularly promising direction as it reduces the size and complexity of deep neural network models while offering a favorable trade-off between size and performance. Even though network pruning has been widely studied and successfully applied in areas such as classification\cite{anwar2017structured}, object detection\cite{liu2024revisiting} and semantic segmentation\cite{he2021cap}, its potential in face detection remains minimally examined. Specifically, only two works in the literature focus on pruning face detection models. Lin et al. \cite{10.1145/3436369.3437415} use the L1 regularization in order to prune the ``least important'' filters in a given layer. However, this approach is potentially suboptimal, as shown in \cite{he2019filter}. Gkrispanis et al. \cite{gkrispanis2024filter} apply Soft Filter Pruning (SFP)\cite{he2018soft} and Filter Pruning via Geometric Median (FPGM)\cite{he2019filter} criterion to selectively prune the most redundant filters, achieving substantial size reduction without significantly sacrificing performance.

This work builds upon \cite{gkrispanis2024filter}, expanding it by optimizing the pruning rate of each layer, instead of pruning the whole network uniformly with the same pruning rate. Specifically, our main contributions are the following:

\begin{itemize}
    \item Utilization of Bayesian optimization for non-uniform pruning across the network in combination with FPGM pruning, a state-of-the-art structured pruning method.
    \item Application of our B-FPGM methodology to EResFD, the smallest well-performing face detector of the literature, to achieve a superior size-to-performance trade-off.
\end{itemize}

This work represents the first application of Bayesian optimization to structured pruning in the literature.

\section{Related Work}
\label{sec:related}

\subsection{Face Detection}
Face detection is a computer vision task that has recently been dominated by deep neural networks. In particular, Convolutional Neural Networks (CNNs) constitute the majority of state-of-the-art face detectors\cite{minaee2021going}.

Many face detection convolutional models that aim for fast inference speed are based on the Single Shot MultiBox Detector (SSD) \cite{liu2016ssd} object detection algorithm, since it follows a single state approach which enhances inference speed while preserving detection performance. Some examples of SSD based face detectors are PyramidBox \cite{tang2018pyramidbox} and RetinaFace \cite{deng2020retinaface}. TinaFace \cite{zhu2020tinaface} approaches face detection as a specialized instance of generic object detection. YOLO5Face \cite{qi2022yolo5face} is derived from the YOLO (You Only Look Once) family of detection models, which exhibit very high inference speeds and thus are widely used in real time applications \cite{jiang2022review}. S3FD \cite{zhang2017s3fd} is a detector model designed for robust face detection across a wide range of scales, with a specific emphasis on detecting smaller faces. The Dual Shot Face Detector (DSFD) \cite{li2019dsfd} proposed a methodology based on the progressive anchor loss, anchor matching and a feature enhancement module. However, many of the aforementioned detection models utilize heavy convolutional backbones, such as ResNet-50 and VGG16. Consequently, even though they may achieve high inference speeds on high-performance hardware, they are impractical for lightweight edge devices.

\subsection{Lightweight Face Detectors}
The use of face detectors in real-time applications and their deployment on edge devices has necessitated the design of lightweight face detector models. RetinaFace \cite{deng2020retinaface} is a widely used lightweight face detector that utilizes MobileNet \cite{howard2017mobilenets} as a backbone. SCRFD \cite{guo2022sample} enhances training with Sample Redistribution and boosts computational efficiency through Computation Redistribution, optimizing the allocation of resources within the model's backbone network. FDLite\cite{aggarwal2024fdlite} utilizes a custom lightweight backbone network (BLite) and two independent multi-task loss functions. Faceboxes \cite{zhang2017faceboxes}, built upon the SSD framework \cite{liu2016ssd}, integrates Rapidly Digested Convolutional Layers (RDCL) and Multiple Scale Convolutional Layers (MSCL) to effectively detect faces of different sizes. Light and Fast Face Detector for edge devices (LFFD) \cite{he2019lffd} optimizes edge device deployment through anchor-free one-stage face detection. Eagle Eye\cite{zhao2019real} utilizes down-sampling and depth-wise convolutions to improve efficiency. The Multi-task Convolutional Neural Network (MTCNN) \cite{8110322} employs Multi-task Cascaded Convolutional Networks with a three-stage cascade architecture. Yunet \cite{wu2023yunet} employs efficient feature extraction and pyramid feature fusion for achieving a low parameter count ($75,856$ parameters), however at the expense of detection accuracy, especially for small face regions. EXTD \cite{yoo2019extd} produces feature maps by repeatedly reusing a shared backbone network to reduce the total number of parameters. The Efficient ResNet Face Detector (EResFD) \cite{jeong2022eresfd} re-evaluates the effectiveness of using standard convolutional blocks as a lightweight backbone for face detection, diverging from the prevalent use of depth-wise convolution layers. By utilizing a heavily channel-pruned ResNet, EResFD achieves superior accuracy compared to similar parameter-sized networks. EResFD is the most compact well-performing face detector in the literature, featuring just $92,208$ parameters.

\subsection{Neural Network Pruning}
Neural network pruning is a technique used to reduce the number of parameters of deep neural networks by removing weights or filters. It aims to reduce memory requirements (and computational complexity) while preserving or even potentially increasing a model's performance. Pruning has gained popularity due to the increasing need for deploying models on edge devices.

Network pruning approaches can be categorized into structured and non-structured methods. Non-structured pruning removes individual parameters, leading to irregular sparsity patterns. Irregular sparsity patterns are challenging to exploit since they require specialized hardware or software for the efficient inference of the pruned models. Conversely, structured pruning techniques remove whole model components, such as filters, producing models that are more straightforward to deploy efficiently. As a result, structured pruning is attracting growing interest in the literature \cite{li2016pruning, Luo_2017_ICCV, anwar2017structured, fang2023depgraph}.

Another way to classify pruning methods is into uniform and non-uniform pruning methods\cite{gong2024fast}. Uniform methods apply a consistent pruning rate across the entire network, whereas non-uniform methods vary the pruning rates for different layers. Some of the most common uniform pruning methods are magnitude-based, such as the $L1$ or $L2$ methods. Those methods prune the weights or filters with the smallest magnitudes \cite{cheng2024survey}. In contrast, non-uniform methods often rely on engineered rules to determine the different pruning rates across the network. For instance, Polyak \& Wolf \cite{polyak2015channel} manually assign higher pruning rates for the earlier layers compared to the later ones. Ma et al. \cite{ma2023llm} cluster the network modules into groups and assign an importance score to each group based on the gradient information from the next token loss. Sun et al.\cite{sun2023simple} calculate weight importance based on the product of its magnitude and the norm of the activation. Liu et al.\cite{liu2021layer} perform layer-wise ranking and apply more aggressive quantization to the least significant network layers. Molchanov et al.\cite{molchanov2019importance} conduct filter-wise sensitivity analysis and iteratively prune the least significant ones. 

Instead of relying on engineered rules, some non-uniform approaches automatically adjust the pruning rate for each network layer. One of the most notable methods in this category is AMC\cite{he2018amc}. AMC utilizes Reinforcement Learning (RL) in order to search for the optimal pruning rate for each layer. Building on AMC, two other works also leverage RL to optimize the pruning rate of each layers. Bencsik et al. \cite{bencsik2022efficient} incorporate a State Predictor Network as a simulated environment rather than validating the pruned model in run time. PURL \cite{gupta2020learning} uses a dense reward scheme, instead of providing a reward only at the end of the episode. However, a limitation of PURL is that it performs non-structured pruning. Additionally, while RL has been successfully applied to various domains, it has several drawbacks, including high training computational complexity, sample inefficiency, nonstationarity, and convergence issues \cite{wang2022deep}. As a result, methods in the literature that automatically adjust a network’s pruning rates remain relatively underdeveloped.

Furthermore, certain methods apply structured network pruning by formulating the pruning process as a mask optimization problem. This approach was initially proposed by Tiwari et al. \cite{tiwari2021chipnet}. Bu et al.\cite{bu2021learning} further extend this concept by proposing a method that jointly prunes and trains the network. Li et al.\cite{li2023differentiable} refine this approach by introducing an efficient end-to-end trainable scheme.
 
While all the previously discussed methods hard-prune the network's parameters, Soft Filter Pruning (SFP) \cite{he2018soft} introduces a novel approach. Instead of fixing the pruned filters' values to zero, SFP allows these filters to be updated during subsequent training steps. 

Gkrispanis et al.\cite{gkrispanis2024filter} successfully utilize SFP within the field of face detection. They utilize a combination of SFP\cite{he2018soft} and FPGM pruning\cite{he2019filter} to iteratively perform soft FPGM pruning and retraining of the face detection network. Subsequently, they apply hard FPGM pruning followed by fine-tuning to obtain the final pruned network. During both the soft and hard pruning, they follow the uniform pruning approach.

\begin{figure}[h!]
\begin{center}
\begin{tabular}{c}
\includegraphics[width=0.96\linewidth]{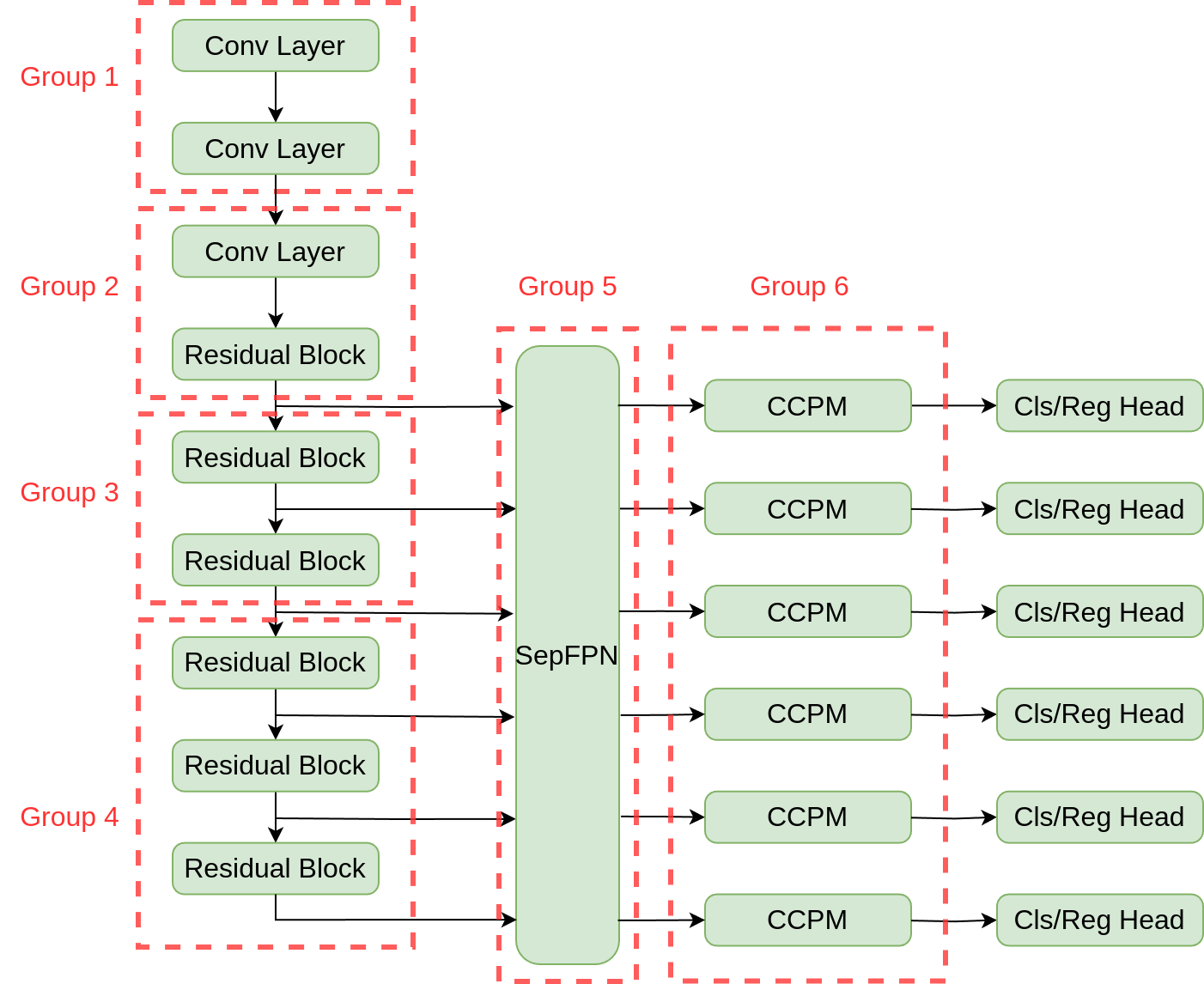} \\
(a) \\
\includegraphics[width=0.8\linewidth]
{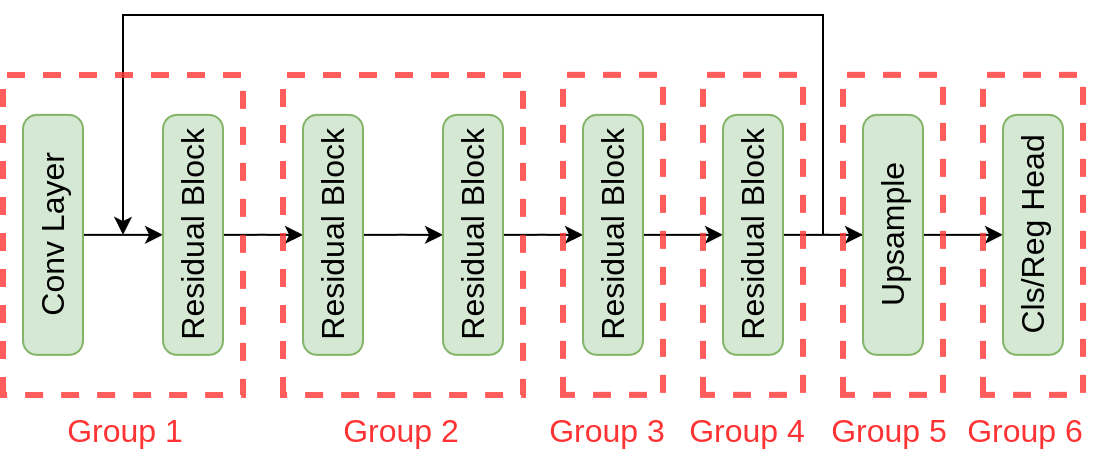} \\
(b) 
\end{tabular}
\end{center}
\caption{Layer grouping of (a) the EResFD model, (b) the EXTD model. Each red rectangle represents a distinct group of layers. SepFPN and CCPM stand for Separated Feature Pyramid Network and Cascade Context Prediction Module respectively. See \cite{jeong2022eresfd} and \cite{yoo2019extd} for a detailed description of the network architectures.}
\label{fig:archs}
\end{figure}

\subsection{Bayesian optimization}
Optimization is a subfield of mathematics that focuses on finding an optimal solution for a specific problem, which is defined by an objective function. Various approaches have been proposed in order to minimize or maximize a given objective function. Bayesian Optimization is one of those approaches. It uses a probabilistic model (e.g. a Gaussian Regression model) to approximate the objective function. The goal is to identify optimal solutions with a minimal number of objective function evaluations \cite{frazier2018tutorial}. Additionally, it stochastically explores the solution parameter space while simultaneously exploiting information from previous evaluations. Thus, it adheres to the exploration-exploitation trade-off, a common challenge  in many optimization and machine learning algorithms \cite{ishii2002control, berger2014exploration}. 

Bayesian optimization has been extensively used in the literature and in practice to tune the values of various parameters in real-world problems, such as tuning the hyperparameter values in machine learning models\cite{cho2020basic}. Hence, its usage has been associated with ``taking the human out of the loop'' \cite{shahriari2015taking}, which is highly desirable.

Bayesian optimization has also been utilized for neural network pruning. Tung et al. \cite{tung2017fine} utilize it to guide the search process for identifying which weights to prune. However, their approach leads to unstructured pruning, making it considerably more challenging to exploit than structured pruning methods.

\section{Proposed Methodology}
\label{sec:proposed}
\subsection{Overall pruning approach}

In this study, we address the challenge of pruning a face detection model to achieve a target sparsity $T$, while preserving as much performance of the original model as possible. Given a neural network $\boldsymbol{\theta}$ pre-trained for $n_{pretrain}$ epochs and a target sparsity $T \in (0,1)$, we divide the network into $N$ layer groups. We then employ Bayesian optimization to identify the optimal pruning rate for each group, denoted as $\boldsymbol{\phi}^*$, which is a vector of size $N$. Subsequently, we apply $\boldsymbol{\phi}^*$ within  the pruning methodology of \cite{gkrispanis2024filter}, instead of applying a uniform pruning approach using the uniform pruning vector $\boldsymbol{\phi}=[T, T,...,T]$. Explicitly, \cite{gkrispanis2024filter} performs soft FPGM pruning and re-training of the face detection network for $n_{SFP}$ epochs. Following that, it hard FPGM prunes the network and fine-tunes it for $n_{finetune}$ epochs. The outcome of this step is the final, pruned face detection model.

Overall, our complete methodology consists of these sequential steps:
\begin{itemize}
    \item Pre-training of the network without pruning.
    \vspace{-2pt}
    \item Dividing the network into groups and Bayesian optimization of each group's pruning rate.
    \vspace{-2pt}
    \item Application of the optimized pruning rates $\boldsymbol{\phi}^*$ within the pruning methodology of \cite{gkrispanis2024filter}.
\end{itemize}

An overview of our proposed methodology is presented in Figure \ref{fig:overview}. We apply our methodology to prune the currently smallest (in number of parameters) well-performing face detector of the literature, EResFD \cite{jeong2022eresfd}, in order to achieve a superior size to performance trade-off. To document the generality of the proposed methodology, a small ablation study with a second small face detector, EXTD \cite{yoo2019extd}, is also reported. 

\subsection{Network Layer grouping}
It is well recognized that in deep learning the effect of individual weights on the network's output varies greatly. Hence, several non-uniform pruning approaches have been proposed in the literature. We also aim to exploit the variance in significance within the layers by grouping them and applying an individual pruning rate to each group, instead of using a uniform pruning rate across the entire network. Since Bayesian optimization will be used to optimize the pruning rates of each group, the grouping is designed with the goal of minimizing the dimensionality of the optimization space, while still preserving sufficient flexibility. An overview of the layer grouping strategy applied to the EResFD and EXTD neural network architectures is presented in Figure \ref{fig:archs}. The $N=6$ groups were established in an ad-hoc manner by examining the architecture diagrams. We chose not to prune the classification and regression heads in EResFD, as they account for only 0.7\% of the network's total parameters; contrarily, we prune them in EXTD as they account for a much larger part of the network. The exact number of parameters contained in each group is presented in Table \ref{table:grooup_params}. In Subsection \ref{ssec:ablations}, we present an ablation study on the effect of adopting different layer groupings on the performance of the pruned EResFD.

\begin{table}[t]
\centering
\small
\begin{tabular}{ l|c || l|c }
 \toprule
 EResFD & \# of Param. & EXTD & \# of Param. \\
 \midrule
Group 1& 1208 & Group 1 & 6850\\
 Group 2& 5856& Group 2 & 36612\\
 Group 3 & 28608&Group 3 & 36482 \\
 Group 4& 33568 & Group 4 & 36482\\
 Group 5& 10802& Group 5 & 24000\\
 Group 6 & 11520&Group 6 & 21926 \\
 
 \bottomrule
\end{tabular}
\caption{The number of parameters in each network layer group.}
\label{table:grooup_params}
\end{table}

\subsection{Bayesian optimization}
After dividing the layers into $N$ groups, we optimize the pruning rate of each group using Bayesian Optimization. This step in the pipeline is conducted after the initial pre-training and, thus, before any pruning. Given the discontinuous nature of the optimization problem, calculating derivatives is not feasible. Consequently, derivative-free optimization, also referred to as black box optimization, is the only viable approach. We chose Bayesian optimization, a derivative-free optimization method, because the objective function is expensive to evaluate and Bayesian optimization is suitable for such problems \cite{frazier2018tutorial}. In contrast, approaches such as Genetic Algorithms require numerous iterations and are therefore impractical for costly fitness functions \cite{katoch2021review}.

Since we need to optimize the pruning rates of each group, the optimization parameter space for the Bayesian optimization is of size $N$, where $N$ represents the number of layer groups. During each iteration $i$ of the optimization, a vector $\boldsymbol{\phi}_i$ is sampled from the parameter space. $\boldsymbol{\phi}_i$ is the vector that contains the pruning rate for each group. The bounds for each element of $\boldsymbol{\phi}_i$ are defined as  $[0, T +bound\_offset]$, allowing the pruning rate for each group to vary from 0\% up to the target pruning rate plus an additional offset. For the first $i_0$ iterations, this vector is sampled randomly. Subsequent iterations use an acquisition function to guide the sampling process. The acquisition function used is the Upper Confidence Bound function (UPC), a widely used and efficient acquisition function\cite{srinivas2009gaussian, snoek2012practical}. Once a vector $\boldsymbol{\phi}_i$ is sampled, the objective function value $f_{\boldsymbol{\theta}}(\boldsymbol{\phi}_i)$ is evaluated. After the evaluation, the result is used to update the posterior probability distribution of the objective function $f_{\boldsymbol{\theta}}$. Upon completion of the optimization process after $I$ iterations, the optimal solution $\boldsymbol{\phi}^*$ is returned. The Bayesian optimization algorithm is presented in Algorithm \ref{alg:bayesian-opt}. 

\begin{algorithm}[t]
\caption{Bayesian optimization}
\label{alg:bayesian-opt}
\begin{algorithmic}
\REQUIRE Initial number of points $i_0$
\REQUIRE Total number of iterations $I$
\REQUIRE Objective function $f$
\STATE Evaluate $f$ at $i_0$ random initial points
\STATE $i \leftarrow i_0$
\WHILE{$i <= I$}
    \STATE Update the posterior probability distribution on objective function $f$ using all available data
    \STATE Sample vector $\boldsymbol{\phi}_i$ using the acquisition function
    \STATE Evaluate $f(\boldsymbol{\phi}_i)$
\ENDWHILE
\STATE \textbf{return} optimal solution $\boldsymbol{\phi}^*$
\end{algorithmic}
\end{algorithm}

\begin{algorithm}[t]
\caption{Overall approach to face detector pruning and training}
\label{alg:fpgm-training-pipeline}
\begin{algorithmic}
\STATE Initialize neural network $\boldsymbol{\theta}$
\STATE train $\boldsymbol{\theta}$ for $n_{pretrain}$ epochs
\STATE Divide the network $\boldsymbol{\theta}$ layers into groups
\STATE Optimize the pruning rate for each group using Bayesian optimization and determine optimal pruning rates $\boldsymbol{\phi}^*$
\FOR{epoch in $[0, n_{SFP}-1]$} 
    \IF{epoch $mod \, 5 = 0$}
        \STATE soft-prune $\boldsymbol{\theta}$ using $\boldsymbol{\phi}^*$
    \ENDIF
    \STATE train $\boldsymbol{\theta}$ for one epoch
\ENDFOR
\STATE Prune $\boldsymbol{\theta}$ using $\boldsymbol{\phi}^*$ and freeze pruned parameters
\STATE Train $\boldsymbol{\theta}$ for $n_{finetune}$ epochs
\end{algorithmic}
\end{algorithm}

The selection of the optimization objective function $f_{\boldsymbol{\theta}}$ is critical. Its value must be as inexpensive as possible to evaluate, as it needs to be assessed in every iteration during the optimization process. The chosen objective function is the validation loss $f_{loss}$ of the network after having performed on it a trial non-iterative soft FPGM pruning and one-epoch training (i.e. just one step of the full pruning process on \cite{gkrispanis2024filter}), supplemented by an additional term $g$ to ensure that the network is pruned approximately at the target pruning rate. If the pruned network sparsity is not within an acceptable range around the target sparsity, the objective function value is set to $penalty\_value$. This indicates that the solution $\boldsymbol{\phi}_i$ sampled by the Bayesian optimization is not satisfactory, and in response to this we conserve computational time by not performing the one epoch of training and loss function evaluation.

Specifically, after sampling a solution vector $\boldsymbol{\phi}_i$, the pre-trained network $\boldsymbol{\theta}$ undergoes soft FPGM pruning (by applying the criterion of Subsection \ref{sec:fpgm}) to generate $\boldsymbol{\theta}_{\boldsymbol{\phi}_i}$, which is then trained for one epoch to obtain $\boldsymbol{\theta}'_{\boldsymbol{\phi}_i}$. The objective function $f_{\boldsymbol{\theta}}$ is formally defined as:
 \begin{equation}
    f_{\boldsymbol{\theta}}({\boldsymbol{\phi}}_i) =  \left\{
\begin{array}{l}
      f_{loss}(\boldsymbol{\theta}'_{\boldsymbol{\phi}_i}) + \lambda \cdot  g(\boldsymbol{\theta}_{\boldsymbol{\phi}_i}, T) \\
      \quad\quad\;\;, \text{if} \;\; T - T^+ \leq S(\boldsymbol{\theta}_{\boldsymbol{\phi}_i}) \leq T + T^+\\
      penalty\_value\quad\quad\quad\quad\quad\quad ,\text{otherwise} \\
\end{array} 
 , \right.
\label{eq:objective_fn}
    \end{equation}

\noindent where $T$ is the target pruning rate, $T^+$ is the acceptable sparsity deviation threshold, $f_{loss}$ is the validation loss function, $\lambda$ is a regularization parameter, $g$ is the sparsity penalty function and $S$ is a function that returns the sparsity of a network. The sparsity penalty function $g$ is defined as:
\begin{equation}
g(\boldsymbol{\theta}_{\boldsymbol{\phi}_i}, T)= 
     \left\{ \begin{array}{ll}
        T - S(\boldsymbol{\theta}_{\boldsymbol{\phi}_i}) & , \text{if} \;\; T > S(\boldsymbol{\theta}_{\boldsymbol{\phi}_i})\\
       0 & ,\text{otherwise} \\
\end{array} 
\right. .
\label{eq:g}
\end{equation}

\noindent By adjusting the value of $\lambda$ we signal to the optimizer the relative importance of deviating from the target sparsity.

It is important to note that this methodology solely optimizes the network layer group pruning rates $\boldsymbol{\phi}^*$ and still requires an additional processing step to prune and optimize the network using these rates.

\subsection{FPGM pruning}
\label{sec:fpgm}
Having determined the optimal group pruning rates $\boldsymbol{\phi}^*$, we proceed with applying to pre-trained network $\boldsymbol{\theta}$ an adaptation of the FPGM pruning algorithm of \cite{gkrispanis2024filter}.

In a convolutional layer of $\boldsymbol{\theta}$, given a set of filters $\boldsymbol{F} = [ \boldsymbol{F_1}, \dots, \boldsymbol{F_M]}$, where, $\boldsymbol{F}_m \in \mathbb{R}^{k \times k \times c}$ is the $m$th filter with spatial size $k \times k$, depth $c$,  $M$ is the total number of filters in the layer and $\boldsymbol{x}$ is a filter in the filter space $\mathbb{R}^{ {k \times k \times c} }$, FPGM first calculates the geometric median of that layer:
 \begin{equation}
    \boldsymbol{x}^{\text{GM}} = \arg\min_{\boldsymbol{x} \in \mathbb{R}^{ {k \times k \times c} }} \sum_{m \in [1, {M}]} \|\boldsymbol{x} - \boldsymbol{F}_{m}\|_2.
    \end{equation}
Thereafter, the filters closest to $\boldsymbol{x}^{\text{GM}}$ are deemed redundant and can be pruned.
After the pre-training of the network $\boldsymbol{\theta}$, we iteratively apply this pruning criterion, using the optimal pruning rates $\boldsymbol{\phi}*$, to soft-prune the filters closest to the geometric median and re-train the network for $n_{SFP}$ epochs. Subsequently, we hard-prune the network and train it for $n_{finetune}$ epochs, following the paradigm of \cite{gkrispanis2024filter}. The overall pruning approach is summarized in Algorithm \ref{alg:fpgm-training-pipeline}.

\begin{table}[t]
\centering
\small
\begin{tabular}{ l|c || l|c }
 \toprule
 Hyperparameter & Value & Hyperparameter & Value\\
 \midrule
$N$& 6 & $I$ & 1000\\
 $n_{pretrain}$& 300& $T^+$ & 0.04\\
 $n_{SFP}$ & 200&$bound\_offset$ & 0.2 \\
 $n_{finetune}$ & 10 & $penalty\_value$ & 100\\
 $i_0$ & 60&$\lambda$ & 5\\
 \bottomrule
\end{tabular}
\caption{Hyperparameter values used.}
\label{tab:hparams}
\end{table}

\begin{table*}[t]
\centering
\small 
\begin{tabular}{lc|ccc|ccc}
\toprule
Method  &$T$& Easy & Medium & Hard & \centering\arraybackslash Actual Sparsity & Actual Sparsity per layer group & \# of Params. \\
\midrule
EResFD (orig.) \cite{jeong2022eresfd} && 0.8902 & 0.8796 & 0.8041 & 0\% & n/a & 92,208\\
EResFD (rep.) \cite{gkrispanis2024filter} && 0.8660 & 0.8555 & 0.7731 & 0\% & n/a & 92,208\\
\midrule
uniform \space FPGM \cite{gkrispanis2024filter} & 10\% & \textbf{0.8728} & \textbf{0.8582} & \textbf{0.7757} & 5.25\% & same as overall sparsity &87,368\\
B-FPGM &10\% & 0.8622 & 0.8506 & 0.7636 & \textbf{10.24\%}& 20.0, 0.0, 7.9, 20.0, 20.0, 17.2\% & \textbf{82,765}\\
\midrule
uniform \space FPGM \cite{gkrispanis2024filter} &20\% & 0.8369 & 0.8201 & 0.7230 & 16.84\% & same as overall sparsity &76,677 \\
B-FPGM &20\% & \textbf{0.8601} & \textbf{0.8477} &\textbf{0.7475} & \textbf{22.27\%} & 0.0, 0.0, 0.0, 40.0, 40.0, 39.8\% & \textbf{71,673}\\

\midrule
uniform \space FPGM \cite{gkrispanis2024filter} &30\% & 0.8311 & 0.8160 & 0.7175 & 24.36\% & same as overall sparsity & 69,746\\
B-FPGM &30\% & \textbf{0.8348} & \textbf{0.8266} & \textbf{0.7231} &  \textbf{31.59\%} & 21.3, 4.5, 30.5, 47.1, 32.1, 44.8\% & \textbf{63,079}\\

\midrule
uniform \space FPGM \cite{gkrispanis2024filter} &40\% & 0.8124 & 0.7952 & 0.6807 &35.95\% & same as overall sparsity & 57,055\\
B-FPGM &40\% & \textbf{0.8227} & \textbf{0.8192} & \textbf{0.7205} & \textbf{40.02\%} & 0.8, 5.2, 47.2, 60.0, 24.4, 46.4\% & \textbf{55,306} \\

\midrule
uniform \space FPGM \cite{gkrispanis2024filter} &50\% & 0.7103 & 0.6830 & 0.5254 & 48.72\% & same as overall sparsity & 47,284\\
B-FPGM &50\% & \textbf{0.7956} &  \textbf{0.7955} & \textbf{0.6993} & \textbf{50.37\%} & 0.0, 0.0, 38.3, 70.0, 56.1, 70.0\% & \textbf{45,578}\\

\midrule
uniform \space FPGM \cite{gkrispanis2024filter} &60\% & 0.5209 & 0.4566 & 0.2936 & 54.05\% & same as overall sparsity & 42,369\\
B-FPGM &60\% & \textbf{0.6974} &  \textbf{0.6937} & \textbf{0.6051} & \textbf{59.87\%} & 8.2, 0.0, 80.0, 80.0, 80.0, 16.4\% & \textbf{37,030}\\

\bottomrule
\end{tabular}
\caption{Comparative results (mAP) on EResFD between uniform FPGM \cite{gkrispanis2024filter} and the proposed B-FPGM. $T$ is the target pruning rate.}
\label{tab:bfpgm-comparsion-eres}
\end{table*}

\begin{table*}[htb]
\centering
\small 
\begin{tabular}{lc|ccc|ccc}
\toprule
Method  &$T$& Easy & Medium & Hard & \centering\arraybackslash Actual Sparsity & Actual Sparsity per layer group & \# of Params. \\
\midrule
EXTD (orig.) \cite{jeong2022eresfd} && 0.9210 & 0.9110 &  0.8560 & 0\% & n/a & 162,352\\
EXTD (rep.) \cite{gkrispanis2024filter} && 0.8961 & 0.8868 & 0.8268 & 0\% & n/a & 162,352\\
\midrule
uniform \space FPGM \cite{gkrispanis2024filter} &20\% & 0.8931 & 0.8789 & 0.7992 & 16.05\% & same as overall sparsity &136,296 \\
B-FPGM &20\% & \textbf{0.9058} & \textbf{0.8905} &\textbf{0.8126} & \textbf{20.3\%} & 31.1, 15.7, 32.4, 21.6, 31.1, 9.8\% & \textbf{129,395}\\
\bottomrule
\end{tabular}
\caption{Comparative results (mAP) on EXTD between uniform FPGM \cite{gkrispanis2024filter} and the proposed B-FPGM. $T$ is the target pruning rate.}
\label{tab:bay-fpgm-comparsion-extd}
\end{table*}

\begin{figure*}[h!] 
   \begin{subfigure}{0.25\textwidth}
       \includegraphics[width=\linewidth]{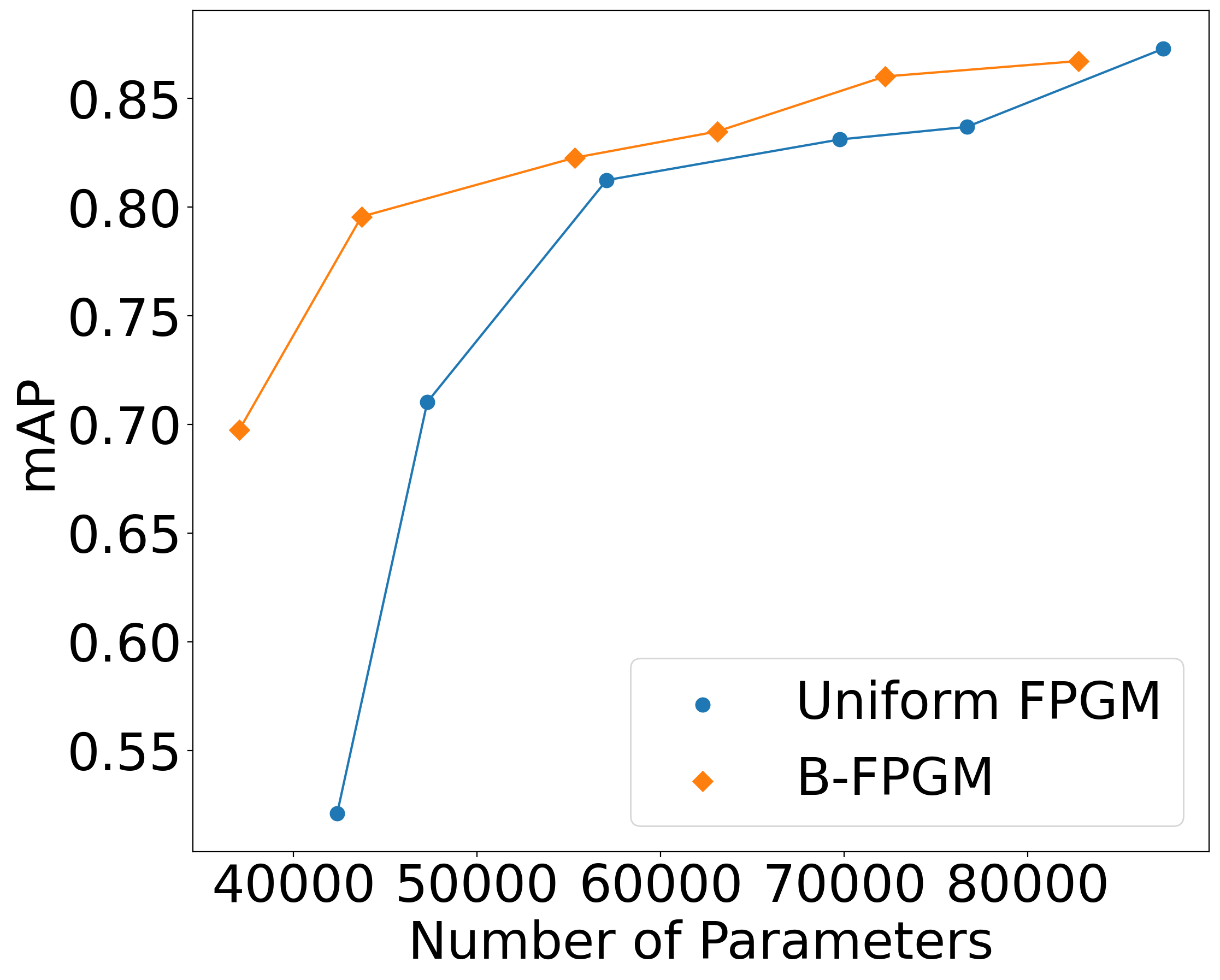}
       \caption{}
       \label{fig:easy_curve}
   \end{subfigure}
\hfill 
   \begin{subfigure}{0.25\textwidth}
       \includegraphics[width=\linewidth]{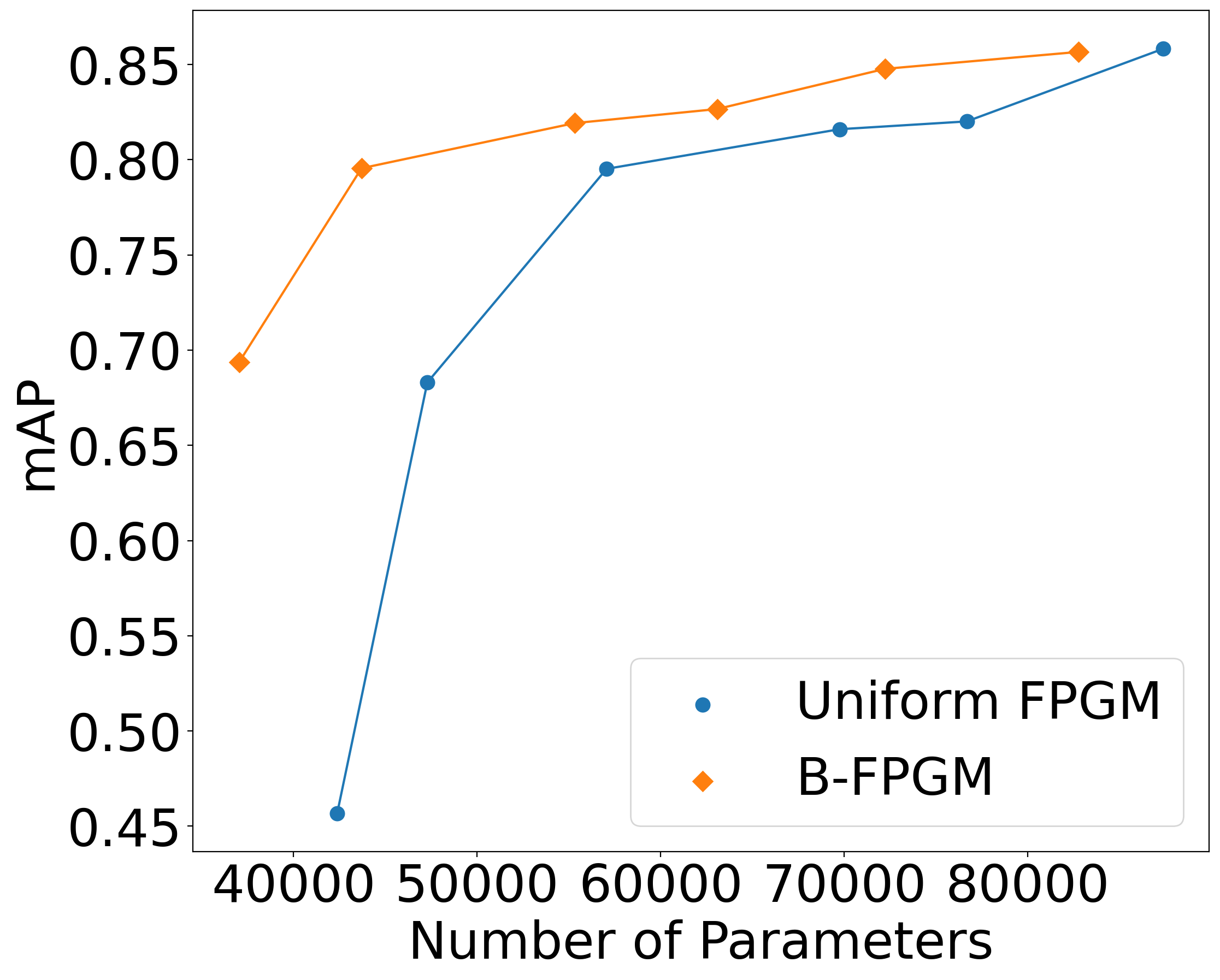}
       \caption{}
       \label{fig:medium_curve}
   \end{subfigure}
\hfill 
   \begin{subfigure}{0.25\textwidth}
       \includegraphics[width=\linewidth]{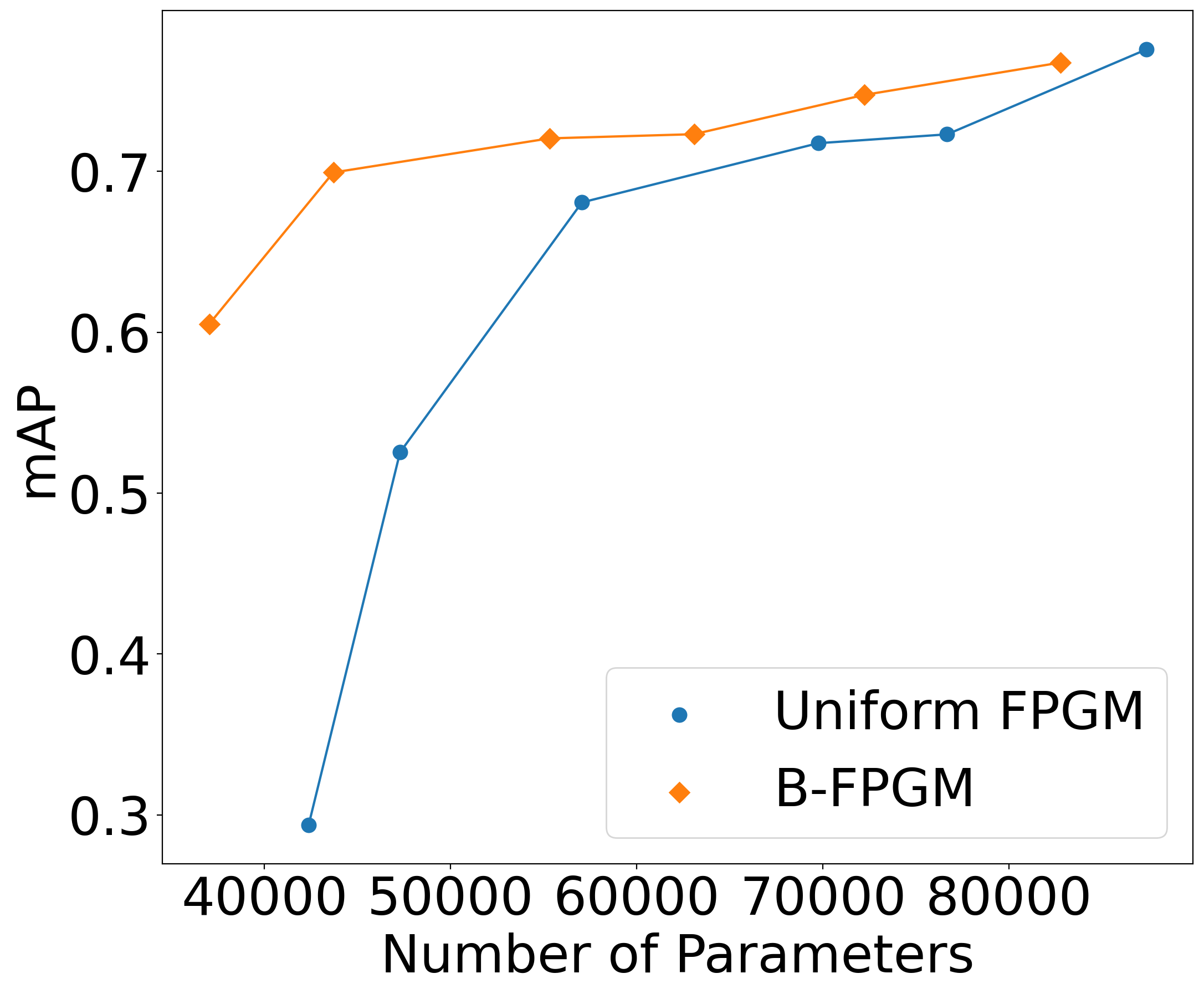}
       \caption{}
       \label{fig:hard_curve}
   \end{subfigure}

   \caption{Comparative results (mAP) on the (a) Easy, (b) Medium, (c) Hard WIDER FACE subsets.}
   \label{fig:image2}
\end{figure*}

\section{Experiments}
\label{sec:experiments}

\subsection{Dataset and Metrics}
\label{ssec:dataset_metric}

The WIDER FACE dataset \cite{yang2016wider}, a widely utilized benchmark in face detection research, was utilized for both training and evaluation in this work. It encompasses 32,203 images that vary in terms of lighting, occlusions, pose and resolution (approx. between 600×1024 and 1300×1024 pixels). Similarly to other works in the literature, e.g.  \cite{jeong2022eresfd, 8110322}, we used the ``validation set'' for evaluation, since the labels for the test set are not publicly available. To use as a validation set (since our objective function requires the validation loss to be calculated), during the Bayesian optimization step only we further split the training set into training and validation subsets, in an 80\%-20\% split.

We utilize Mean Average Precision (mAP@0.5) as our evaluation metric, a widely used metric both in the face detection literature and in combination with the WIDER FACE dataset. Our test set is divided into three subsets based on difficulty, Easy (1146 images), Medium (1079 images), Hard (1001 images), as in \cite{yang2016wider}. We calculated each model's mAP@0.5 separately for each subset. 

In our experiments we used the same hyperparameter values as \cite{gkrispanis2024filter}; these are summarized in Table \ref{tab:hparams}.

\begin{table}[htb]
\centering
\small
\begin{tabular}{ cc|ccc|c}
 \toprule
 \textit{N} & \textit{T} & Easy &  Medium & Hard & Sparsity\\
 \midrule
 4  & 20\%  & 0.8093 & 0.8035 & 0.6860 &  \textbf{24.45\%} \\
 6  &20\%  &  \textbf{0.8601} &  \textbf{0.8477} &  \textbf{0.7475} &22.27\% \\
 9  &20\%  & 0.8390 & 0.8264 &0.7105 &23.99\% \\
 \midrule
 4  &50\%  & 0.7887 & 0.7519 & 0.6457 & \textbf{50.75}\% \\
 6  &50\%  & \textbf{0.7956} &  \textbf{0.7955} &  \textbf{0.6993} &  50.37\% \\
 9  &50\%  & 0.7896 & 0.7931 & 0.6855 & 49.69\% \\
 \bottomrule
\end{tabular}
\caption{mAP of B-FPGM on EResFD, on WIDER FACE (Easy, Medium, Hard subsets), for different network layer groupings. \textit{N} is the number of layer groups and \textit{T} is the target pruning rate.}
\label{tab:group_ablation}
\end{table}

\begin{table}[htb]
\centering
\small
\begin{tabular}{ l|c}
 \toprule
 Subset & Mean mAP $\pm$ standard deviation  \\
 \midrule
 Easy & $0.8565 \pm 0.0052$ \\
 Medium & $0.8454 \pm 0.0044$ \\
 Hard& $0.7490 \pm 0.0066$ \\
 \bottomrule
\end{tabular}
\caption{Mean mAP $\pm$ standard deviation of B-FPGM on EResFD across five runs, using different random seeds, for $T$=20\%.}
\label{tab:random_seeds}
\end{table}

\subsection{Results}
\label{ssec:results}
We experimented with pruning rates ranging from 10\% to 60\% and initially compared the proposed B-FPGM pruning to uniform FPGM pruning \cite{gkrispanis2024filter}. The results in Table \ref{tab:bfpgm-comparsion-eres} and Fig. \ref{fig:image2} show that the B-FPGM-pruned models consistently achieve a superior size to performance trade-off compared to their uniform-FPGM-pruned counterparts. The actual sparsity per layer group of the B-FPGM-pruned models, reported in Table \ref{tab:bfpgm-comparsion-eres}, varies considerably between layer groups, and tends to be higher for Groups 3 to 6 of EResFD. 

\begin{figure*}[t]
    \centering
    \includegraphics[width=0.9\linewidth]{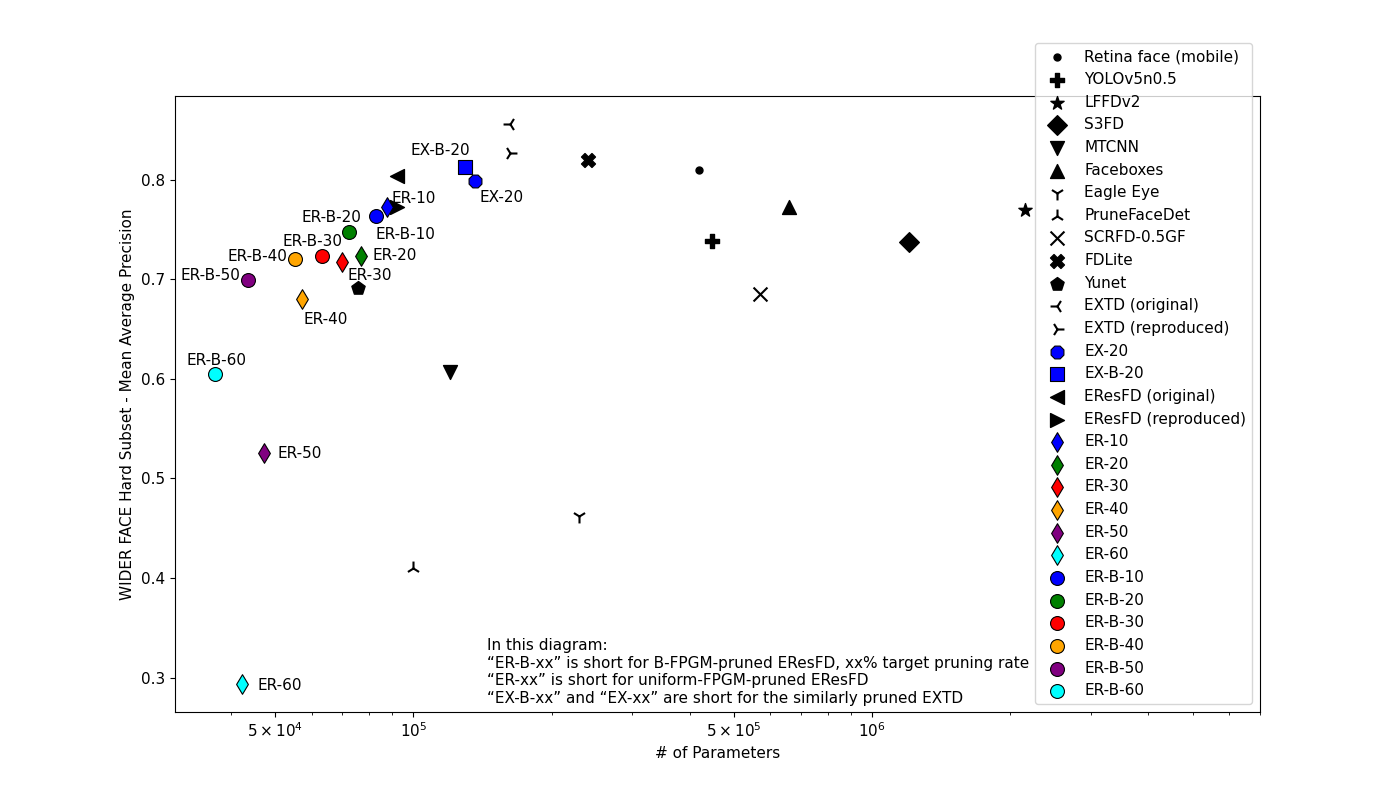}
    \caption{Model size and mAP across different face detector models on the Hard subset of WIDER FACE.}
    \label{fig:model_comparison}
\end{figure*}

Figure \ref{fig:model_comparison} provides a comparative analysis of B-FPGM-pruned EResFD models against SoA lightweight face detectors, on the WIDER FACE hard subset. We compare against lightweight models, namely RetinaFace (with MobileNet as backbone)\cite{deng2020retinaface}, Yolo5Face \cite{qi2022yolo5face}, EagleEye \cite{s19092158}, LFFD \cite{he2019lffd}, FDLite\cite{aggarwal2024fdlite}, S3FD \cite{zhang2017s3fd}, MTCNN \cite{8110322}, FaceBoxes \cite{zhang2017faceboxes}, SCRFD \cite{guo2022sample}, PruneFaceDet \cite{10.1145/3436369.3437415}, Yunet \cite{wu2023yunet}, and the original EXTD \cite{yoo2019extd} and EResFD \cite{jeong2022eresfd} as well as their pruned versions using uniform FPGM pruning. This figure illustrates that our B-FPGM approach applied to the EResFD detector provides a superior size-to-performance trade-off compared to SoA lightweight face detectors.

\begin{figure}[h!]
\begin{center}
\begin{tabular}{cc}
EResFD 
& B-FPGM 50\% pruning \\
\includegraphics[width=0.44\linewidth]{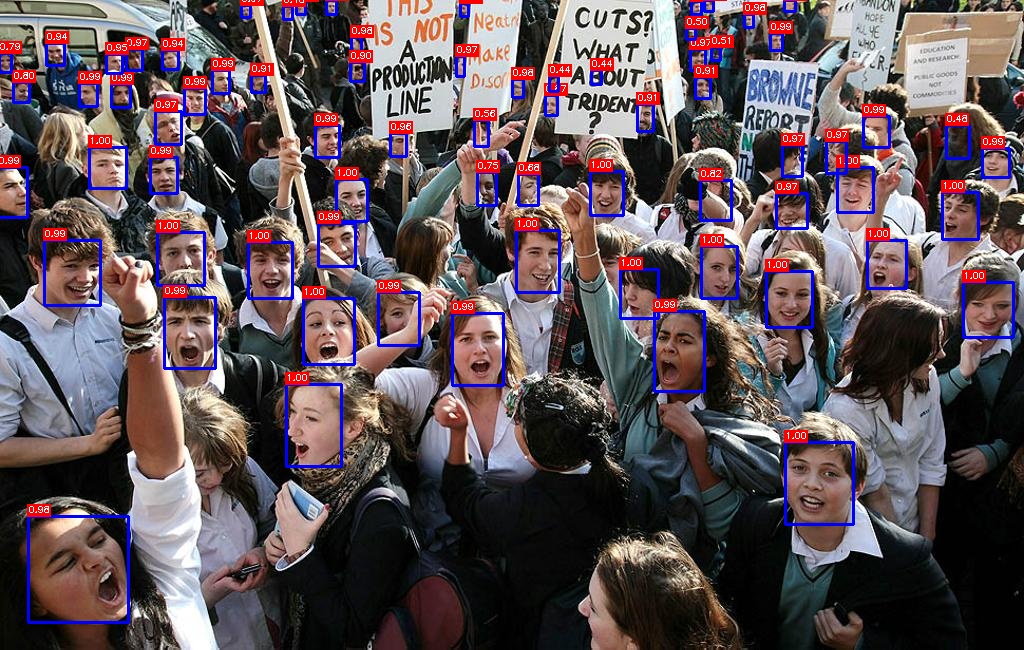} &
\includegraphics[width=0.44\linewidth]{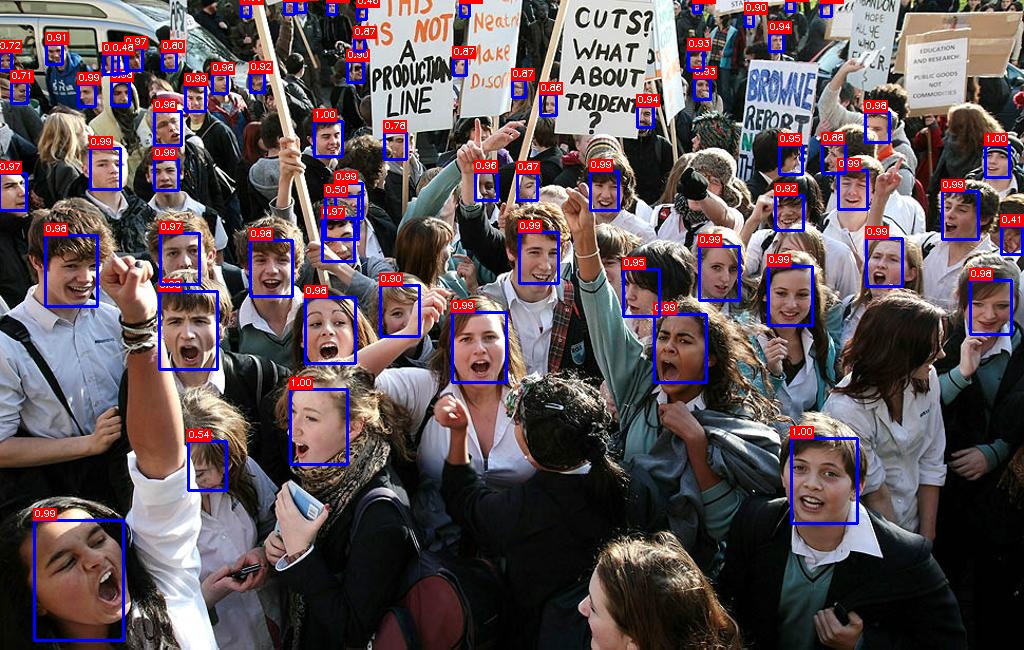} \\
\includegraphics[width=0.44\linewidth]{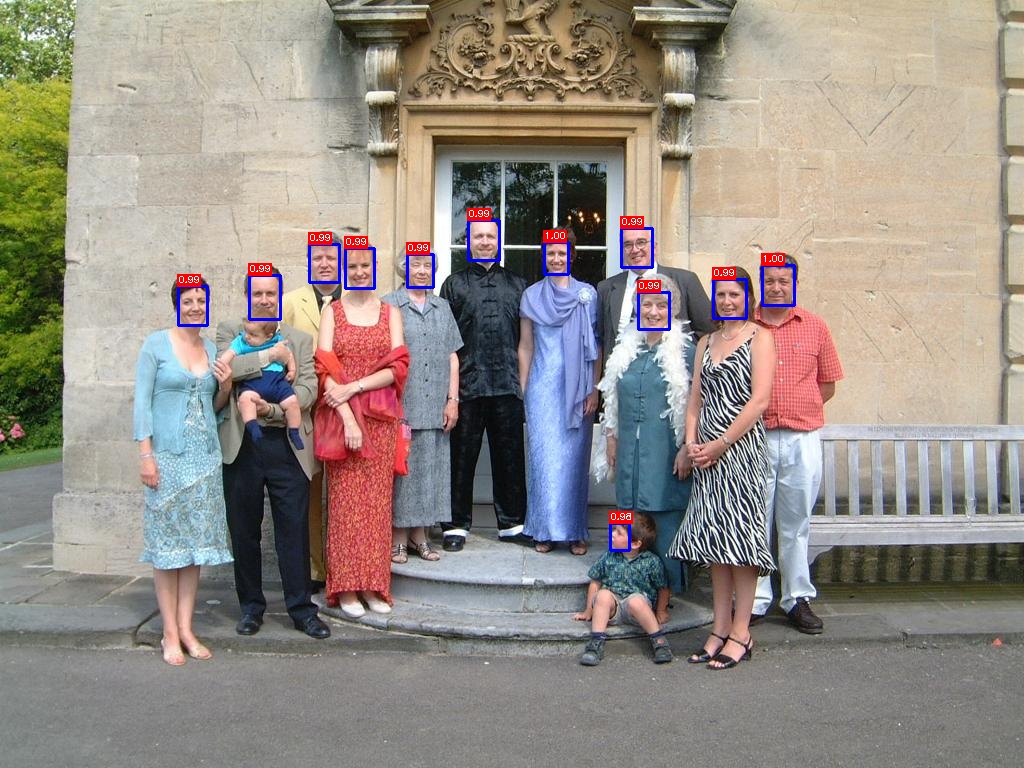} &
\includegraphics[width=0.44\linewidth]{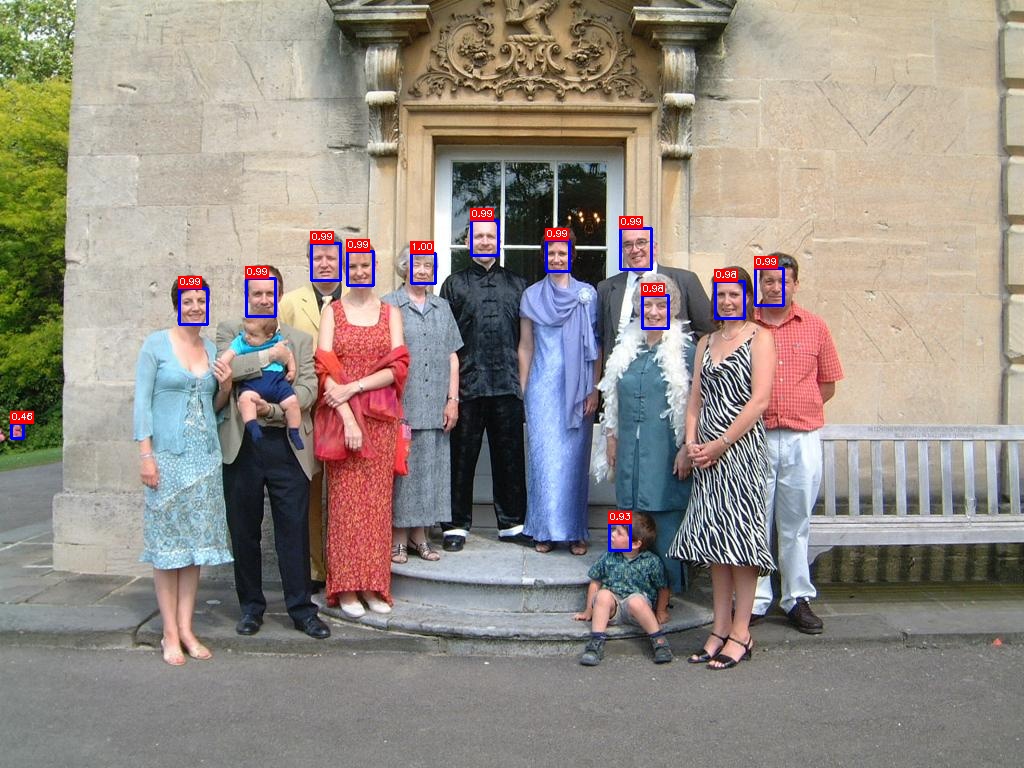} \\
\end{tabular}
\end{center}
\caption{Examples of face detection results using the original EResFD model and its 50\% B-FPGM-pruned variant.}
\label{fig:inference}
\end{figure}

Finally, Fig. \ref{fig:inference} presents a couple of face detection examples, using the original EResFD model alongside its 50\% B-FPGM-pruned variant. These examples indicate that the network's performance does not significantly decline as sparsity increases even up to 50\%: the pruned model still demonstrates the ability to accurately detect small faces.

\subsection{Ablations}
\label{ssec:ablations}

We conducted a small ablation study to examine the applicability of B-FPGM pruning to a different network architecture; for this, we pruned another small face detector, EXTD. The results in Table \ref{tab:bay-fpgm-comparsion-extd} are consistent with those obtained on EResFD, regarding the superiority of B-FPGM over uniform FPGM. As expected, the actual sparsity per layer group of the B-FPGM-pruned EXTD model differs from those observed on EResFD.

In another ablation we examined the impact of different layer groupings on the performance of EResFD. Specifically, we experimented with $N=4$ and $N=9$ layer groups ($N=4$: merging together Groups 1-2 and Groups 3-4 of the grouping shown in Fig. \ref{fig:archs}(a) for $N=6$; $N=9$: further splitting each of Groups 3 and 4 of Fig. \ref{fig:archs}(a) according to the stages of the EResFD model, i.e. to two and three groups, respectively). The comparative results on Table \ref{tab:group_ablation} illustrate that $N=6$ strikes a good balance. Additionally, given the probabilistic nature of Bayesian optimization, to assess the robustness of our algorithm to stochasticity, we repeated the 20\% pruning experiment five times using different random seeds. The results (Table \ref{tab:random_seeds}) document that our algorithm exhibits minimal variance across runs.

\section{Conclusions}
\label{sec:conclusions}

We presented a new network pruning approach that leverages FPGM pruning, Soft Filter Pruning and Bayesian optimization, in order to optimize the pruning rates of each part of the network and achieve superior trade-off between network size and performance. Applied to the small EResFD face detector, our approach consistently resulted in pruned face detectors that surpass the SoA performance for the same level of sparsity. Especially at higher pruning rates, the improvement over the SoA is particularly prominent; up to 50\% of the EResFD network's parameters can be pruned with only a moderate impact on performance. An ablation study on another small face detection network, EXTD, demonstrated the generality of the proposed pruning methodology.

{\small
\bibliographystyle{ieee_fullname}
\bibliography{Bibliography}
}

\end{document}